\documentclass[12pt]{article}
\usepackage[margin=1in]{geometry}
\usepackage{amsmath,amsfonts}
\usepackage{array}
\usepackage[caption=false,font=normalsize,labelfont=sf,textfont=sf]{subfig}
\usepackage{textcomp}
\usepackage{hyperref}
\hypersetup{hidelinks}
\usepackage{url}
\usepackage{verbatim}
\usepackage{graphicx}
\usepackage{booktabs}
\usepackage{algorithm}
\usepackage{algpseudocode}
\DeclareMathOperator*{\argmin}{arg\,min}
\def\BibTeX{{\rm B\kern-.05em{\sc i\kern-.025em b}\kern-.08em
    T\kern-.1667em\lower.7ex\hbox{E}\kern-.125emX}}
\usepackage{setspace}
\providecommand{\IEEEPARstart}[2]{#1#2}
\newcommand{\PaperAppendix}[1]{\appendix\section*{Appendix: #1}}

\newcommand{\blindfootnote}[1]{%
  \begingroup
  \renewcommand{\thefootnote}{}\footnote{#1}%
  \addtocounter{footnote}{-1}%
  \endgroup
}

\begin{document}

\title{A Peer-Relative Representation Learning Framework for Energy
Inefficiency Identification in Mobile Network Sites}
\author{Eliud~Nyakweba~Koto\thanks{E.~N.~Koto is with the African Institute for Mathematical Sciences (AIMS), Cape Town, South Africa (e-mail: eliud@aims.ac.za).},
Jaco~du~Toit\thanks{J.~du~Toit is with Technology Strategy Planning Architecture \& Assurance. Vodacom Group Limited, South Africa, and the Department of Electrical and Electronic Engineering, Stellenbosch University, South Africa, (e-mail: jacowp357@gmail.com)},
Adham~Stoltz\thanks{A.~Stoltz is with Technology Strategy Planning Architecture \& Assurance, Vodacom Group Limited, South Africa, and School of Computer Science and Applied Mathematics, University of the Witwatersrand, South Africa, (e-mail: Adham.Stoltz@vodacom.co.za).},
and~Johan~du~Preez\thanks{J.~du~Preez is with the Department of Electrical and Electronic Engineering, Stellenbosch University, Stellenbosch, South Africa (e-mail: jadupreez@gmail.com).}
}
\date{}

\maketitle
\blindfootnote{This work was supported in part by a Google DeepMind Scholarship at AIMS South Africa.}

\begin{abstract}
Energy consumption is one of the largest operational expenditure items for mobile network operators, yet site-level energy inefficiencies such as faulty cooling controllers, idle radio equipment, and parasitic auxiliary loads often remain undetected because no ground-truth inefficiency labels
exist and historical measurements may already contain embedded inefficiencies. This study proposes an unsupervised peer-relative approach based on the premise that sites with similar structural and operational characteristics should exhibit comparable energy consumption. To capture these relationships, a novel energy-aware Minimum Distortion Embedding (MDE) formulation is introduced that extends the standard MDE objective with an energy-based repulsion mechanism. This encourages sites with anomalously high energy consumption relative to comparable peers to become displaced from their local neighbourhoods in the embedding space. The resulting low-dimensional representation simultaneously preserves structural similarity and encodes energy-related deviations, enabling the identification of potentially inefficient sites through peer-relative comparison. The derived anomaly scores provide a practical mechanism for prioritising field investigations, allowing mobile network operators to focus engineering resources on sites most likely to yield energy savings. Experimental results demonstrate that the proposed approach outperforms conventional anomaly detection baselines and provides a robust foundation for large-scale energy-efficiency optimisation in mobile networks.

\end{abstract}

\noindent\textbf{Keywords---} Anomaly detection, energy efficiency, green communications, minimum-distortion embedding, mobile networks, representation learning, unsupervised learning.

\section{Introduction}
\begin{figure}[!ht]
\centering
\includegraphics[width=0.7\columnwidth]{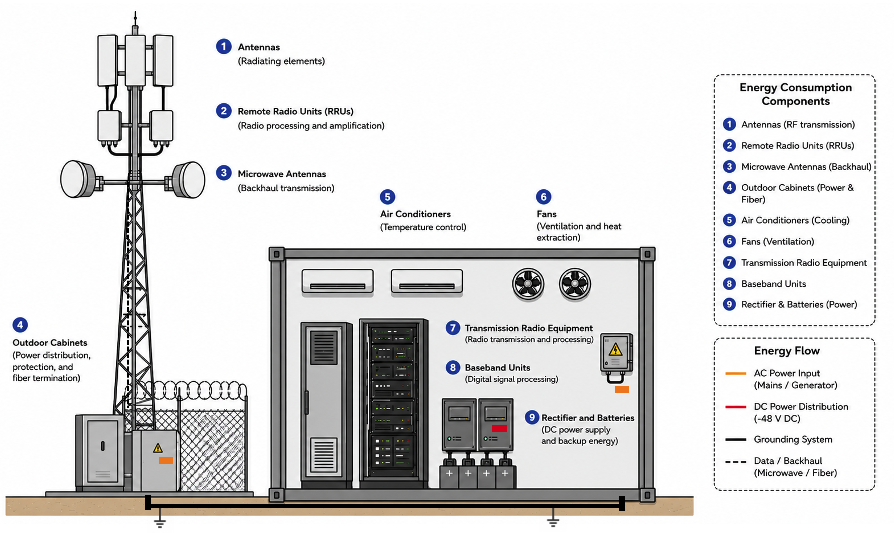}
\caption{High-level component layout of a typical mobile network site. Radio and signal processing equipment provide the communication functions, while supporting subsystems such as cooling and power conversion contribute a large proportion of the site's overall energy consumption.}
\label{fig:site_layout}
\end{figure}

\IEEEPARstart{T}{he} rapid growth of mobile communications has driven a substantial expansion of the infrastructure operated by mobile network operators (MNOs). Modern networks comprise thousands of geographically distributed sites, each a complex cyber-physical system combining radio access network (RAN) equipment including antennas, remote radio units (RRUs), baseband units
(BBUs), and transmission hardware, with supporting infrastructure such as rectifiers, backup batteries, power conversion equipment, remote management systems, and thermal management systems (Fig.~\ref{fig:site_layout}). The cumulative energy demand
of these sites constitutes a major component of MNO operational expenditure and a growing sustainability concern as electricity prices rise, carbon-reduction commitments tighten, and network densification continues  \cite{ngmn2024green,lopezperez2022survey}. Because base station sites dominate network energy consumption, even modest site-level efficiency improvements translate into substantial financial
and environmental savings when scaled across a national network.

A significant share of site energy is consumed by supporting subsystems rather than by the radios themselves. Thermal management is a prominent example: radio, power, and electronic components generate heat inside equipment enclosures, and cooling systems must keep them within safe operating
temperatures \cite{zhang2022cooling}. Measurements at an operational site in South Africa showed that cooling alone accounted for approximately $40\%$ of total site energy consumption while a single air conditioner was active \cite{su2024cooling}. Faults in these subsystems, such as a
controller that fails to switch off an air conditioner, a degraded rectifier, parasitic auxiliary loads, or radio equipment drawing near-peak power under low utilisation, can waste energy persistently without causing service interruptions, and therefore remain unnoticed for long periods.

Traditional approaches to reducing site energy consumption have primarily focused on engineering interventions, including designing and validating improved physical site configurations in laboratory or test environments before rolling them out across the network, as well as optimising parameters on active network equipment. While these approaches have delivered substantial improvements, they assume that similar optimisations can be broadly applied across sites. In practice, however, every mobile network site exhibits a unique combination of characteristics, including its equipment configuration, physical layout, environmental exposure, local weather conditions, traffic demand, and operational history. These factors interact to produce highly site-specific energy consumption behaviour, making it difficult to directly compare measured electricity consumption with billing records or identify genuine opportunities for optimisation from absolute energy usage alone.

Detecting such inefficiencies from operational data is difficult for three reasons. First, there are typically no labels indicating whether a site operates efficiently: monthly energy readings blend normal operation with possible abnormal behaviour, ruling out conventional supervised learning
\cite{himeur2021ai}. Second, models trained to predict historical consumption may absorb long-standing inefficiencies into the learned baseline: a site whose air conditioning has run continuously for months appears \emph{normal} in its own history, so a regression model learns the inflated
usage as expected consumption \cite{song2022noisy}. Third,
energy consumption is meaningful only in context. A site may consume more energy because it is genuinely inefficient, but also because it is larger,
carries more traffic, or hosts a more complex configuration
\cite{tan2022energy,piovesan2022ml}; fixed thresholds and population-global outlier detection therefore conflate structural heterogeneity with inefficient operation.

This paper addresses these challenges by treating energy inefficiency as a peer-relative pattern: a site becomes a candidate for inspection when its energy behaviour is inconsistent with that of structurally comparable
peers. We instantiate this idea through minimum-distortion embedding (MDE) \cite{agrawal2021mde}, a general framework for learning low-dimensional representations from pairwise relationships. Our formulation embeds sites so  that structural peers with consistent energy behaviour remain close, while an
energy-aware repulsion term pushes sites with unusually high consumption away from their local peer groups. The resulting displacement provides an anomaly score that ranks sites for field investigation, and the ranking can further be
distilled into lightweight supervised classifiers through pseudo-labels.

The main contributions of this paper are as follows.
\begin{itemize}
\item We propose an energy-aware MDE formulation that integrates structural similarity and local energy deviation in a single push--pull distortion objective. To the best of our knowledge, this is the first application of distortion-based representation learning to site-level energy inefficiency detection in mobile networks.
\item We derive an embedding-relative displacement score that quantifies how far each site sits from its structural peer group, normalised by the intrinsic spread of that group, and we show how the score supports pseudo-label generation for downstream supervised models.
\item We construct a controlled evaluation environment grounded in operational data from $5{,}372$ live mobile network sites, with group-specific baseline energy models and four realistic classes of injected inefficiency (overload, cooling overhead, idle-RF load, and non-RAN parasitic load), enabling quantitative evaluation despite the absence of real inefficiency labels.
\item We present a comprehensive empirical study against unsupervised
detectors, supervised residual-based methods, and a privileged physics
reference across contamination rates from $1\%$ to $99\%$, together with a representation-level analysis showing that the learned embedding itself improves standard detectors, and a teacher--student experiment demonstrating a practical deployment pathway.
\end{itemize}

To support reproducibility, the full implementation and experimental code are publicly available at~\cite{koto2026peerrelative}.

The remainder of this paper is organised as follows.
Section~\ref{sec:related} reviews related work.
Section~\ref{sec:framework} presents the proposed energy-aware embedding framework. Section~\ref{sec:dataset} describes the controlled evaluation methodology, and Section~\ref{sec:setup} the experimental setup.
Section~\ref{sec:results} reports and discusses the results, and
Section~\ref{sec:conclusion} concludes the paper.

\section{Related Work}
\label{sec:related}

\subsection{Energy Consumption Modelling in Mobile Networks}

RANs account for a substantial share of the operational energy demand of mobile telecommunication systems, with base station sites identified as the main contributor to overall network energy consumption \cite{ngmn2024green}. Research on network energy efficiency has consequently concentrated on modelling RAN power demand and on optimisation mechanisms such as sleep modes, lean carrier design, resource allocation, and traffic-aware management \cite{tan2022energy}. Machine learning has been applied to energy modelling and network optimisation, including regression models, analytical power consumption frameworks, and deep learning methods for estimating site-level power demand \cite{piovesan2022ml}.

Site-level energy consumption is shaped by a combination of technical and operational factors including radio equipment, traffic load, transmission configuration, site architecture, cooling, and auxiliary infrastructure \cite{zhang2022cooling,tan2022energy}, so energy use cannot be interpreted independently of site context. A high-consumption site may be operating normally given its structural complexity and demand, while a lower-consumption site could still be inefficient relative to comparable peers. Inefficiency
detection therefore differs from energy prediction: the objective is not to estimate consumption, but to identify consumption that is unusual for structurally similar sites. Comparatively little work addresses this site-level question, particularly in the absence of ground-truth inefficiency labels, the common constraint in operational environments.

\subsection{Learning Under Noisy Targets}

A recurring finding in the noisy-label literature is that corrupted training signals degrade model reliability in ways not always visible from standard metrics, for instance, models trained on incorrectly labelled examples may memorise noise rather than learn generalisable patterns \cite{song2022noisy}.
Although this literature is usually framed around classification, the same concern applies here because observed energy consumption is only an imperfect proxy for the underlying efficiency state of a site. A site that has operated
inefficiently for an extended period appears normal in historical data because its inefficiency has become part of the observed baseline; a model trained to reproduce that baseline encodes the inefficiency as expected behaviour.
Noisy-label methods such as noise-tolerant loss functions \cite{zhou2023asymmetric} and co-teaching \cite{han2018coteaching} address corrupted supervision in classification, but they assume an observed class-label setting with a defined label-noise process, whereas no verified labels exist in our setting at all. This motivates a departure from purely predictive modelling.

\subsection{Anomaly Detection and Peer-Relative Comparison}
Anomaly detection identifies observations that deviate from expected behaviour, with approaches ranging from supervised to unsupervised depending on label availability, and spanning statistical, clustering-based, density-based, nearest-neighbour, and reconstruction-based techniques \cite{chandola2009survey}. This work operates in the label-free (unsupervised) setting, consistent with the absence of ground-truth inefficiency labels in operational networks. Isolation Forest \cite{liu2008isolation} and local outlier factor (LOF) \cite{breunig2000lof} illustrate two contrasting notions
of abnormality: isolability from the broader population versus deviation from the local neighbourhood density. When abnormality is estimated only relative to the full population, the resulting score may reflect structural differences between sites rather than inefficient operation. Peer-relative comparison addresses this by interpreting each site within a local neighbourhood of structurally similar peers, consistent with local methods such as LOF, but this framing remains underdeveloped for site-level energy inefficiency
detection in mobile network infrastructure. Graph-based representation methods have been used for unsupervised anomaly detection in other domains \cite{zhang2023graph}, and MDE \cite{agrawal2021mde} provides a flexible framework for encoding pairwise relationships into embeddings; however, existing formulations do not incorporate an energy-consistency signal into the embedding objective.

\subsection{Positioning of This Work}

The literature addresses mobile network energy primarily through predictive modelling and network-level optimisation, with limited attention to whether a site consumes more than would be expected from structurally comparable peers. This distinction matters because historical data may already contain inefficiencies, causing predictive models to treat inefficient behaviour as normal, while global anomaly detectors may mistake structural complexity for abnormal consumption. This paper occupies that gap: it treats energy inefficiency as a peer-relative inconsistency and encodes the inconsistency directly into a representation learning objective, providing an unsupervised route to candidate inefficient sites when verified labels are unavailable.

\section{Proposed Peer-Relative Energy-Aware Embedding Framework}
\label{sec:framework}

\subsection{Problem Formulation}
\label{sec:problem}

Let $i = 1,\dots,N$ index the set of mobile network sites. Each site is represented by a feature vector $\mathbf{x}_i \in \mathbb{R}^d$, partitioned as
\begin{equation}
\mathbf{x}_i = (\mathbf{s}_i, e_i),
\label{eq:partition}
\end{equation}
where $\mathbf{s}_i \in \mathbb{R}^{d_s}$ collects structural attributes and $e_i \in \mathbb{R}$ is the observed monthly energy consumption, so that $d = d_s + 1$. The separation in \eqref{eq:partition} allows structural features to define \emph{which} sites should be compared, while the energy measurement is used to assess \emph{whether} a site behaves unusually relative to those peers.

The goal is to learn a low-dimensional representation $\mathbf{z}_i \in \mathbb{R}^p$ in which structurally similar sites with consistent energy behaviour remain close, while sites whose consumption is unusually high relative to their structural peers are displaced from their local group. Because verified labels of efficient and inefficient operation are unavailable, the problem is treated as fully unsupervised, and the displacement provides the inefficiency score.

\subsection{Structural Graph Construction}
\label{sec:graph}

Structural peer relationships are represented by a $k$-nearest-neighbour (kNN) graph built from the structural vectors $\mathbf{s}_i$. Categorical attributes are encoded and numerical attributes scaled so that features with larger numeric ranges do not dominate the distance computation. The framework uses structural neighbourhoods at three scales, all computed from the same structural distance: a graph neighbourhood of size $k_{\mathrm{graph}}$ for embedding construction, a baseline neighbourhood of size $k_{\mathrm{base}}$ for local energy comparison (Section~\ref{sec:objective}), and a scoring neighbourhood of size $k_{\mathrm{score}}$ for displacement scoring (Section~\ref{sec:scoring}). For $k \in \{k_{\mathrm{graph}}, k_{\mathrm{base}}, k_{\mathrm{score}}\}$, the set of the $k$ nearest structural peers of site $i$ is denoted $\mathcal{N}_k(i)$. The graph neighbourhood is computed over the full population, whereas the baseline and scoring neighbourhoods are restricted to sites sharing the vendor, sharing status, and mast group of site $i$, so that energy comparisons are made only among directly comparable configurations. Each site is connected to its $k_{\mathrm{graph}}$ nearest structural peers. The resulting graph is
\begin{equation}
G^{(s)} = \bigl(V, E^{(s)}, W^{(s)}\bigr),
\label{eq:graph}
\end{equation}
where $V$ is the node set, $E^{(s)}$ the structural edge set, and $W^{(s)}$ the edge weights. An edge is included when at least one endpoint selects the other as one of its predefined number of nearest structural neighbours. Concretely, an edge is assigned a weight of $1$ if only one of nodes $i$ and $j$ identifies the other as a nearest neighbour, and $2$ if the relationship is reciprocal. Thus, $W^{(s)}$ encodes neighbour agreement rather than structural distance, which is used only to determine the edge set $E^{(s)}$. This is the standard graph construction used in the MDE framework \cite{agrawal2021mde}.

\subsection{Energy-Aware MDE Objective}
\label{sec:objective}

\subsubsection{Standard MDE}

Let $\mathbf{Z} \in \mathbb{R}^{N \times p}$ denote the embedding matrix whose $i$th row is $\mathbf{z}_i$, and let
\begin{equation}
d_{ij} = \lVert \mathbf{z}_i - \mathbf{z}_j \rVert_2
\label{eq:dist}
\end{equation}
be the embedded distance of a connected pair $(i,j) \in E^{(s)}$. Standard MDE minimises a distortion objective over the graph edges,
\begin{equation}
\mathcal{E}_{\mathrm{MDE}}(\mathbf{Z})
= \sum_{(i,j) \in E^{(s)}} w^{(s)}_{ij}\, \phi(d_{ij}),
\label{eq:mde}
\end{equation}
where $w^{(s)}_{ij}$ is the structural weight and $\phi(\cdot)$ a pairwise distortion function \cite{agrawal2021mde}. The graph specifies which pairs enter the objective; the distortion function determines how their embedded distances are penalised.

\subsubsection{Energy-Aware Weight Adjustment}

The structural graph defines comparable peers but is blind to energy behaviour. We therefore modify the weights of structurally connected pairs before optimisation, reducing attraction between peers when one or both consume unusually high energy relative to their local baseline, and reversing the relationship into repulsion when the deviation is sufficiently large.

Using the baseline neighbourhood $\mathcal{N}_{k_{\mathrm{base}}}(i)$, the local energy baseline of site $i$ is the $q$th percentile of its neighbours' consumption,
\begin{equation}
b_i = \max\!\bigl(P_q\bigl(\{ e_j : j \in \mathcal{N}_{k_{\mathrm{base}}}(i) \}\bigr),\, 1\bigr),
\label{eq:baseline}
\end{equation}
where $P_q(\cdot)$ denotes the percentile operator. The percentile level $q$ controls how conservative the baseline is: lower values anchor it to the lower-energy portion of the neighbourhood, while higher values move it toward the neighbourhood's central tendency. The floor of $1$\,kWh prevents unstable
ratios for very small baselines. The log-ratio deviation of site $i$ is
\begin{equation}
\delta_i = \log\!\left(\frac{e_i}{b_i}\right),
\label{eq:logratio}
\end{equation}
with positive values indicating excess consumption. For each structural edge
$(i,j) \in E^{(s)}$, the edge-level excess is
\begin{equation}
r_{ij} = \max(\delta_i, \delta_j, 0),
\label{eq:excess}
\end{equation}
which captures the larger positive deviation at either endpoint and ignores
edges whose endpoints are both at or below their baselines. The energy-aware
weight is then
\begin{equation}
w_{ij} = w^{(s)}_{ij} - \beta\, r_{ij},
\label{eq:weight}
\end{equation}
where $\beta > 0$ controls the strength of the adjustment. When $r_{ij}$ is small, $w_{ij} \approx w^{(s)}_{ij}$ and the pair remains attractive; when $r_{ij}$ is large, $w_{ij}$ may become negative, converting the interaction from attraction to repulsion. This balance forms the push--pull mechanism at the core of the proposed formulation.

\subsubsection{Push--Pull Objective}

The energy-aware structural edges are combined with a set of sampled dissimilar pairs $E^{(-)}$ that provide additional repulsive scaffolding by keeping selected non-neighbouring sites separated. $E^{(-)}$ contains $\mu\,\lvert E^{(s)} \rvert$ pairs sampled uniformly at random among non-adjacent site pairs, where $\mu$ is a sampling multiplier. With
$E^{(*)} = E^{(s)} \cup E^{(-)}$, the proposed objective is
\begin{equation}
\mathcal{E}_{\beta}(\mathbf{Z})
= \sum_{(i,j) \in E^{(s)}} w_{ij}\, \phi_{+}(d_{ij})
+ \sum_{(i,j) \in E^{(-)}} w_{-}\, \phi_{-}(d_{ij}),
\label{eq:pushpull}
\end{equation}
where $w_{-} < 0$ is a fixed weight for dissimilar pairs, and the attractive
and repulsive penalties are
\begin{equation}
\phi_{+}(d) = \log(1 + d), \qquad \phi_{-}(d) = \log(d).
\label{eq:penalties}
\end{equation}
The embedding is obtained by minimising \eqref{eq:pushpull} over the coordinates,
\begin{equation}
\mathbf{Z}^{*} = \argmin_{\mathbf{Z} \in \mathbb{R}^{N \times p}}
\mathcal{E}_{\beta}(\mathbf{Z}),
\label{eq:opt}
\end{equation}
solved with the first-order numerical routines provided by the PyMDE package \cite{agrawal2021mde}, which iteratively update the coordinates along descent directions of the objective.

\subsection{Embedding-Relative Anomaly Scoring}
\label{sec:scoring}

Displacement from structural peers in the learned embedding serves as the inefficiency signal. For site $i$, the mean embedded distance to its scoring neighbourhood $\mathcal{N}_{k_{\mathrm{score}}}(i)$ is
\begin{equation}
D_i = \frac{1}{\lvert \mathcal{N}_{k_{\mathrm{score}}}(i) \rvert}
\sum_{j \in \mathcal{N}_{k_{\mathrm{score}}}(i)} \lVert \mathbf{z}_i - \mathbf{z}_j \rVert_2 .
\label{eq:Di}
\end{equation}
Since embedded neighbourhoods differ in natural spread, $D_i$ is normalised by
the average pairwise distance among the neighbours themselves,
\begin{equation}
S_i = \frac{1}{\lvert \mathcal{P}_i \rvert}
\sum_{(u,v) \in \mathcal{P}_i} \lVert \mathbf{z}_u - \mathbf{z}_v \rVert_2 ,
\label{eq:Wi}
\end{equation}
where $\mathcal{P}_i$ is the set of neighbour pairs within
$\mathcal{N}_{k_{\mathrm{score}}}(i)$. The peer-relative anomaly score is
\begin{equation}
A_i = \frac{D_i}{S_i + \varepsilon},
\label{eq:score}
\end{equation}
with $\varepsilon > 0$ a small constant preventing division by zero in
degenerate neighbourhoods. A large $A_i$ indicates that site $i$ lies far from
its structural peer group relative to the intrinsic spread of that group.
Sites are ranked by $A_i$, with higher scores indicating stronger evidence of
potential energy inefficiency.

\subsection{Pseudo-Label Distillation}
\label{sec:pseudo}

The anomaly ranking can be converted into supervisory targets for downstream models. Let $\psi \in (0,1)$ denote an assumed anomaly proportion and $\tau_\psi$ the score threshold corresponding to the top-$\psi$ fraction of ranked sites. The pseudo-label of site $i$ is
\begin{equation}
\hat{y}_i =
\begin{cases}
1, & A_i \ge \tau_\psi, \\
0, & \text{otherwise}.
\end{cases}
\label{eq:pseudo}
\end{equation}
The resulting binary pseudo-labelled dataset treats strongly displaced sites as candidate inefficient examples. Because the labels derive from the embedding geometry rather than external annotation, they transfer the anomaly structure  discovered by the framework to conventional classifiers.

Algorithm~\ref{alg:pipeline} summarises the complete pipeline.

\begin{algorithm}[!ht]
\caption{Peer-Relative Energy-Aware Embedding and Scoring}
\label{alg:pipeline}

\begin{algorithmic}[1]
\Require Site features $\{(\mathbf{s}_i,e_i)\}_{i=1}^{N}$ (Eq.~\eqref{eq:partition}); neighbourhood sizes $k_{\mathrm{graph}}$, $k_{\mathrm{base}}$, $k_{\mathrm{score}}$; percentile $q$; repulsion strength $\beta$; dissimilar-pair weight $w_{-}$ and sampling multiplier $\mu$; embedding dimension $p$; scoring constant $\varepsilon$; pseudo-label fraction $\psi$

\Ensure Anomaly scores $\{A_i\}$ (Eq.~\eqref{eq:score}); ranked site list; pseudo-labels $\{\hat{y}_i\}$ (Eq.~\eqref{eq:pseudo})

\State Construct the structural kNN graph $G^{(s)}=(V,E^{(s)},W^{(s)})$ (Eq.~\eqref{eq:graph})

\For{each site $i$}
    \State Compute the local baseline $b_i$ (Eq.~\eqref{eq:baseline}).
    \State Compute the log-ratio deviation $\delta_i$ (Eq.~\eqref{eq:logratio}).
\EndFor

\For{each structural edge $(i,j)\in E^{(s)}$}
    \State Compute the edge excess $r_{ij}$ (Eq.~\eqref{eq:excess})
    \State Update the edge weight $w_{ij}$ (Eq.~\eqref{eq:weight})
\EndFor

\State Sample dissimilar pairs $E^{(-)}$.
\State Form the push--pull objective (Eq.~\eqref{eq:pushpull}) with penalties defined in Eq.~\eqref{eq:penalties}.
\State Solve the optimisation problem (Eq.~\eqref{eq:opt}) to obtain the embedding $\mathbf{Z}^{*}$.

\For{each site $i$}
    \State Compute the mean neighbour displacement $D_i$ (Eq.~\eqref{eq:Di})
    \State Compute the neighbourhood spread $S_i$ (Eq.~\eqref{eq:Wi})
    \State Compute the anomaly score $A_i$ (Eq.~\eqref{eq:score})
\EndFor

\State Rank sites in descending order of $A_i$.
\State Generate pseudo-labels $\hat{y}_i$ (Eq.~\eqref{eq:pseudo})
\State Optionally train a downstream classifier using $\{(\mathbf{x}_i,\hat{y}_i)\}$.

\end{algorithmic}
\end{algorithm}

\subsection{Complexity}

Graph construction, dominated by the kNN search, costs
$\mathcal{O}(N \log N)$ with tree- or index-based search in the processed structural space; baseline and edge-weight computation are linear in the number of edges, $\mathcal{O}(Nk_{\mathrm{graph}})$. Each iteration of the first-order MDE solver evaluates distances and gradients over $\lvert E^{(*)} \rvert$ edges, costing $\mathcal{O}\bigl(\lvert E^{(*)} \rvert\, p\bigr)$ per iteration \cite{agrawal2021mde}, and the scoring step is $\mathcal{O}(N k_{\mathrm{score}}^2)$ in the worst case due to the pairwise neighbour spread in \eqref{eq:Wi}. The pipeline scales to national-network populations (tens of thousands of sites) on commodity hardware. 
 
\section{Controlled Evaluation Methodology}
\label{sec:dataset}

Verified inefficiency labels do not exist in operational data, so detection quality cannot be measured directly on live networks. We therefore construct a controlled evaluation environment that preserves the structural and energy characteristics of a real network while providing known injected inefficiencies as ground truth. The injected labels are used only for evaluation and are never used during graph construction, embedding optimisation, or scoring, preserving the unsupervised nature of the method.

\subsection{Operational Reference Data}
\label{sec:refdata}

The reference dataset covers $5{,}372$ unique live mobile network sites of a national operator, each represented by one cross-sectional observation drawn from the period January 2024 to January 2026 (most observations from January 2026).
Each record contains a site identifier, observation month, RAN-sharing
indicator, RAN vendor, mast type, total radio cell count, non-RAN equipment count, packet-switched traffic volume, and measured monthly energy consumption (kWh), which serves as the energy outcome.

Defining a full site structure as the combination of vendor, sharing status, mast type, cell count, and non-RAN equipment count yields $1{,}336$ unique structures, of which $587$ occur more than once. The data therefore contain both repeated comparable configurations, which are necessary for meaningful peer groups, and substantial heterogeneity. Exploratory analysis shows that consumption differs systematically across vendor and sharing groups (shared sites consume more than standalone sites, and vendor~A sites more than vendor~B sites) and increases with cell count, motivating group-specific baseline energy models rather than a single global model.

\subsection{Synthetic Population Generation}
\label{sec:synthetic}

Synthetic sites are sampled with replacement from the empirical joint
distribution of structural attributes in the reference data, preserving the operational composition of the network. Traffic is assigned from reference sites with matching structural profiles, with small random perturbations to avoid exact duplication. Baseline energy is assigned by group-specific linear models fitted to the reference data: sites are grouped by vendor, sharing status, and mast group (a coarser grouping of the recorded mast types into tower, disguised, rooftop, pole, and other), and within group $g$ the expected monthly consumption of site $i$ is
\begin{equation}
\hat{e}_i = \lambda_g + \alpha_g c_i + \gamma_g n_i,
\label{eq:baselinemodel}
\end{equation}
where $c_i$ is the cell count, $n_i$ the non-RAN equipment count, and
$\lambda_g$, $\alpha_g$, $\gamma_g$ group-specific parameters. Traffic is deliberately excluded from the baseline model: expected consumption in the reference data is driven primarily by structural configuration, and traffic influences simulated energy only through the idle-RF injection mechanism (Section~\ref{sec:injection}). The residual baselines in Section~\ref{sec:setup} nevertheless receive traffic as a regressor, so the comparison is not biased in favour of the proposed framework. Natural
operational variability is modelled multiplicatively,
\begin{equation}
e^{(0)}_i = \hat{e}_i \exp(\epsilon_i), \qquad
\epsilon_i \sim \mathcal{N}\bigl(0, \sigma^2_{\log,i}\bigr),
\label{eq:noise}
\end{equation}
where the log-scale noise level
$\sigma_{\log,i} \sim \mathcal{U}(0.02, 0.04)$ is sampled independently per site, so that energy remains positive and larger sites exhibit larger absolute deviations. At this stage all sites are treated as efficient. The reference configuration generates $N = 5{,}000$ synthetic sites with expected energy consumption ranging from approximately $500$ to $9{,}000$\,kWh (median $3{,}654$\,kWh). Vendor~A accounts for $54\%$ of sites, with cell counts ranging from $2$ to $40$ (median ${\approx}23$) and non-RAN equipment counts reaching up to $17$.

\subsection{Controlled Inefficiency Injection}
\label{sec:injection}

A fraction $\rho$ of sites (the \emph{contamination rate}) is selected
uniformly at random and assigned elevated consumption through one of four mechanisms that reflect realistic sources of excess energy observed in operational networks (Table~\ref{tab:injection}). The mechanisms produce heterogeneous deviation patterns rather than a uniform uplift. Using the noisy baseline $e^{(0)}_i$, the final simulated metered energy is
\begin{equation}
e_i =
\begin{cases}
e^{(0)}_i, & \text{site $i$ efficient}, \\
\mathcal{I}_{t_i}\bigl(e^{(0)}_i, \boldsymbol{\theta}_i\bigr), & \text{site $i$ inefficient},
\end{cases}
\label{eq:injected}
\end{equation}
where $t_i \in \{1,2,3,4\}$ is the assigned injection type, $\boldsymbol{\theta}_i$ the site attributes used by the type-specific perturbation, and $\mathcal{I}_{t_i}(\cdot)$ the injection function. The full parameterisation is given in the Appendix. At the reference setting ($N = 5{,}000$, $\rho = 10\%$), $500$ sites are injected, balanced across the four types.

\begin{table}[h]
\caption{Controlled Inefficiency Injection Types}
\label{tab:injection}
\centering
\footnotesize
\begin{tabular}{@{}lll@{}}
\toprule
Type & Source & Interpretation \\
\midrule
1 & Overload & Excess energy from unusually high site load \\
2 & Cooling & Additive overhead from increased cooling demand \\
3 & Idle-RF & High radio energy draw under low utilisation \\
4 & Non-RAN & Parasitic load from auxiliary infrastructure \\
\bottomrule
\end{tabular}
\end{table}

\begin{figure}[!ht]
\centering
\includegraphics[width=0.8\columnwidth]{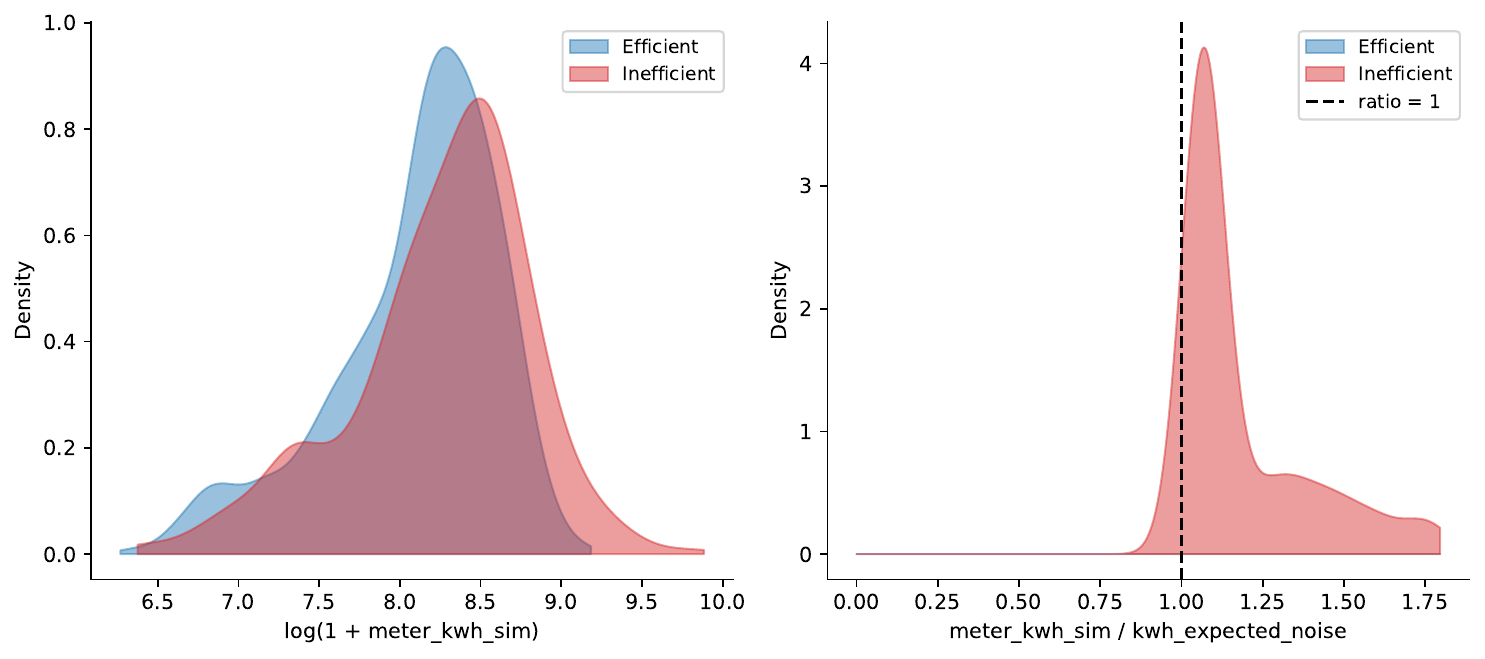}
\caption{Controlled inefficiency patterns in the synthetic dataset at the reference setting: simulated energy distributions on the log scale (left) and the ratio of simulated metered energy to the noisy baseline energy (right) for efficient and injected inefficient sites.}
\label{fig:injection}
\end{figure}

Fig.~\ref{fig:injection} characterises the resulting evaluation setting. Injected sites generally exhibit higher simulated consumption, with the ratio of simulated metered energy to the noisy baseline concentrated near one for efficient sites and shifted towards higher values for injected sites. The blue (efficient) density is not visible because it is concentrated within an extremely narrow region around a ratio of one, making it too narrow to be resolved at the plotting resolution. Although the two populations are well separated in terms of this ratio, the detection problem remains challenging because the anomaly detection methods are evaluated without access to this privileged quantity and must instead infer inefficiency from the available structural and traffic features.

\section{Experimental Setup}
\label{sec:setup}

\subsection{Configuration of the Proposed Framework}

The proposed framework is unsupervised and transductive: the MDE embedding is fitted jointly on all $5{,}000$ synthetic sites without using the injected inefficiency labels, so hyperparameters are selected by partitioning the evaluation labels rather than the features. Using a fixed random seed, the site indices are split into equally sized stratified validation and test partitions at the reference contamination rate of $10\%$; validation labels are used only for selection, and the test partition is held out until the final evaluation.

The three neighbourhood sizes defined in Section~\ref{sec:graph}, for graph construction, local energy comparison \eqref{eq:baseline}, and displacement scoring, serve different purposes and were tuned independently by one-factor-at-a-time grid searches, yielding $k_{\mathrm{graph}} = 300$, $k_{\mathrm{base}} = 10$, and $k_{\mathrm{score}} = 50$. The remaining parameters were fixed a priori: the baseline percentile $q = 35$ as a sub-median reference that limits the influence of inefficient peers, a traffic down-weighting factor of $0.05$ in graph construction, embedding dimension $p = 4$, dissimilar-pair sampling multiplier $\mu = 4$, and dissimilar-pair weight $w_{-} = -2.0$.

\begin{figure}[h]
\centering
\includegraphics[width=0.7\columnwidth]{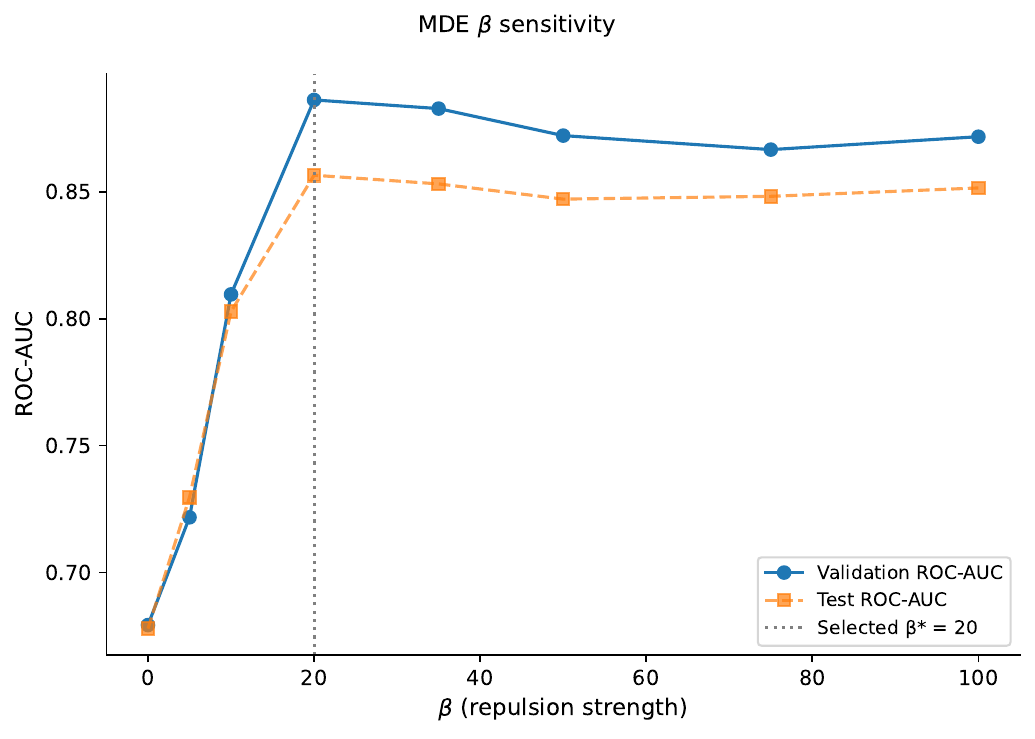}
\caption{Validation-based selection of the repulsion strength $\beta$. Validation and test ROC--AUC are shown across the candidate values, with all remaining hyperparameters held fixed. The vertical dashed line indicates the selected value, $\beta = 20$.}
\label{fig:beta_sensitivity}
\end{figure}

The energy-aware repulsion strength $\beta$, the principal hyperparameter introduced by the framework, was selected by grid search over $\beta \in \{0, 5, 10, 20, 35, 50, 75, 100\}$ with the graph, dissimilar edges, and seeds held fixed; $\beta = 20$ maximises validation ROC-AUC (Fig.~\ref{fig:beta_sensitivity}). The sweep doubles as an ablation and sensitivity analysis: at $\beta = 0$ the purely structural embedding attains a ROC-AUC of only $0.68$, while test ROC-AUC remains within $0.847$--$0.857$ for all $\beta \in [20, 100]$, so the energy-aware term drives detection quality and requires no fine-tuning. The ROC-AUC values in Fig.~\ref{fig:beta_sensitivity} are computed on the half-population validation and test partitions defined above; they are therefore not directly comparable to the full-population results reported in Section~\ref{sec:results}.

\subsection{Experiments and Baselines}

\subsubsection{Experiment 1: Unsupervised Feature-Space Comparison}
The proposed score is compared with standard unsupervised detectors applied to the original site features: Isolation Forest \cite{liu2008isolation}, LOF \cite{breunig2000lof}, a Gaussian mixture model (GMM) scored by negative likelihood, and an autoencoder (AE) scored by reconstruction error \cite{hinton2006ae}. The contamination rate is swept over $\rho \in \{1, 5, 10, 15, 20, 25, 35, 45, 55, 65, 75, 85, 95, 99\}\%$, with all methods evaluated against the same injected labels at each rate.

\subsubsection{Experiment 2: Supervised Residual Comparison}
The proposed unsupervised score is compared with residual-based detectors that predict simulated metered energy from the site features and rank sites by positive residuals: ordinary least squares (LR), robust Huber regression \cite{huber1964robust}, and a random forest (RF) regressor \cite{breiman2001rf}. A \emph{physics residual} reference, which scores sites by the ratio of observed to simulator-expected energy, is included as a privileged upper bound because the simulator's expected energy is not available in deployment.

\subsubsection{Experiment 3: Effect of the Embedding on Standard Detectors}
At the reference contamination rate of $10\%$, Isolation Forest, LOF, GMM, and the AE are each applied in two settings, the original feature space and the learned MDE embedding, to assess whether the energy-aware representation itself improves detection, independently of the proposed displacement score.

\subsubsection{Experiment 4: Teacher--Student Distillation}
The dataset is split $70/30$ with stratification ($3{,}500$ training and $1{,}500$ test sites, preserving the $10\%$ anomaly rate). Pseudo-labels are assigned to the top-ranked $\psi = 10\%$ of training sites by the proposed score, and logistic regression, RF, and XGBoost students are trained on the original features and evaluated on the held-out test set.

\subsection{Evaluation Metrics}

Ranking quality is measured by ROC-AUC, the probability that a randomly chosen injected site outranks a randomly chosen efficient site, and by the area under the precision--recall curve (PR-AUC), computed as average precision, which summarises ranking behaviour under class imbalance \cite{davis2006pr}. Inspection-oriented performance is measured by Precision@$K$ with $K$ equal to the top $10\%$ of ranked sites, reflecting the practical setting in which only the highest-ranked sites are selected for field investigation. Because $K$ matches the number of injected sites at the reference contamination rate of $10\%$, Precision@$K$ and Recall@$K$ coincide in this setting, so only the former is reported.

\section{Results and Discussion}
\label{sec:results}

\subsection{Comparison With Unsupervised Feature-Space Methods}

\begin{figure}[!ht]
\centering
\includegraphics[width=0.9\columnwidth]{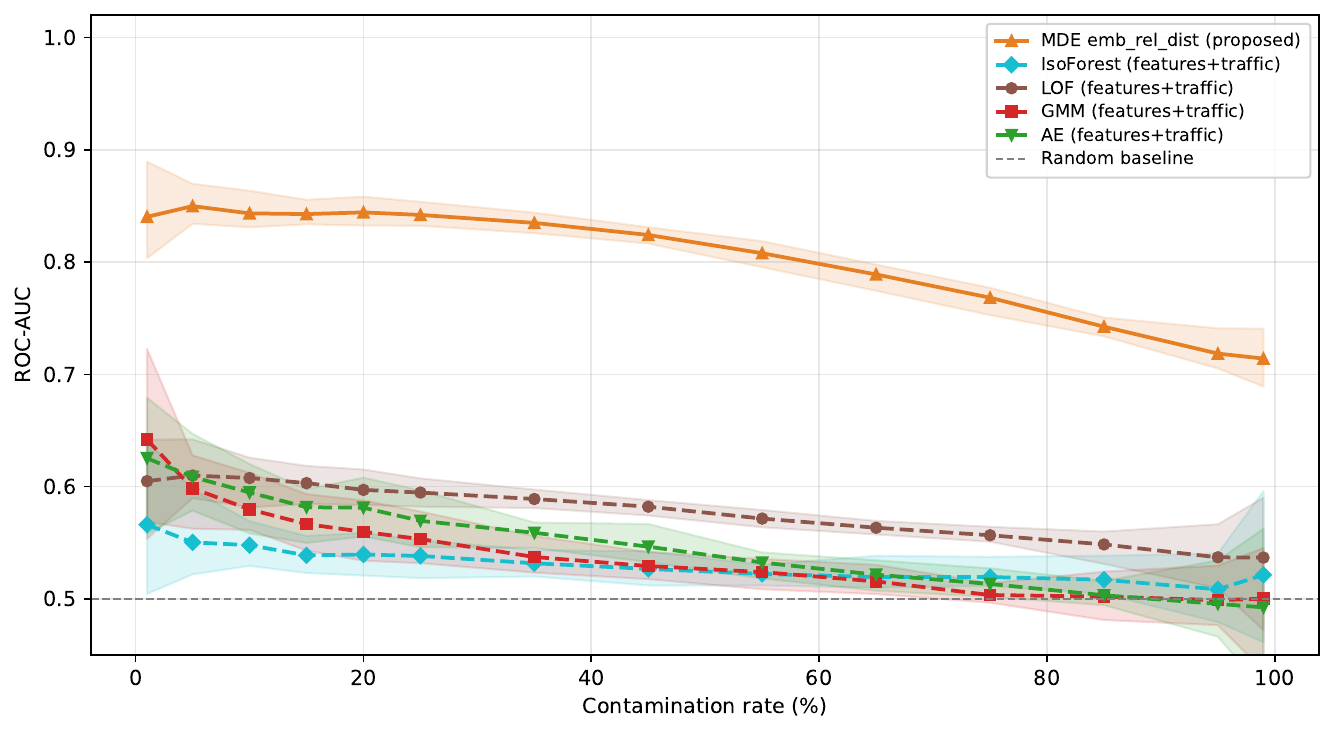}
\caption{ROC-AUC of the proposed energy-aware MDE score and unsupervised feature-space baselines across contamination rates. The proposed method consistently achieves the highest detection performance and remains robust as contamination increases.}
\label{fig:roc_unsup}
\end{figure}

Fig.~\ref{fig:roc_unsup} reports ROC-AUC across the contamination sweep. The proposed score achieves the strongest ranking performance at every rate: it attains $0.84$ at $\rho = 1\%$, degrades only gradually as contamination increases, and remains at $0.71$ even at $\rho = 99\%$. The feature-space baselines remain far below: LOF is the strongest conventional detector (${\approx}0.60$ at low rates), while Isolation Forest, GMM, and the AE remain close to random for most of the sweep. Injected inefficient sites are thus not
generic outliers in the raw feature space, indicating that the signal emerges only when structural similarity and peer-relative energy inconsistency are combined, as the proposed distortion objective does explicitly.

\subsection{Comparison With Supervised Residual Methods}

\begin{figure}[!ht]
\centering
\includegraphics[width=0.9\columnwidth]{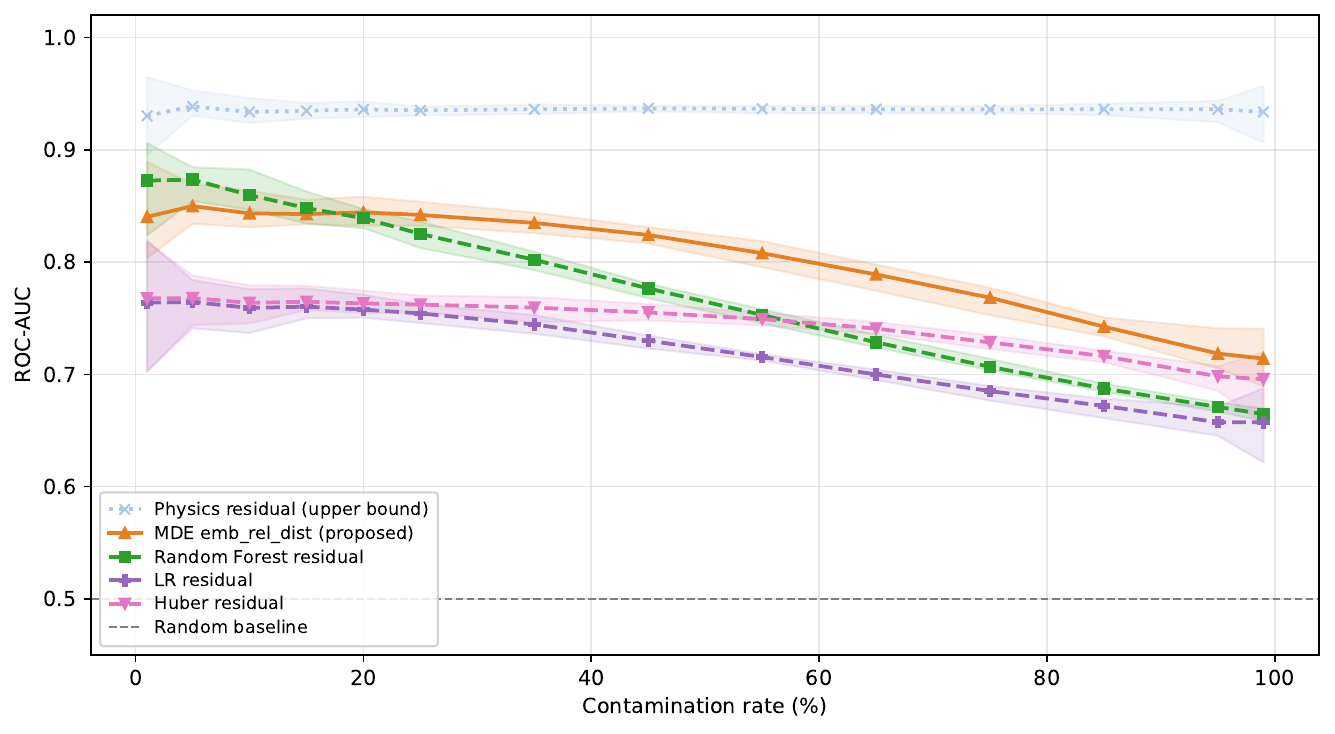}
\caption{ROC-AUC of the proposed energy-aware MDE score and residual-based methods across the contamination-rate sweep. The proposed method consistently outperforms the practical supervised baselines, while the physics residual is shown only as a privileged upper bound.}
\label{fig:roc_sup}
\end{figure}

Fig.~\ref{fig:roc_sup} shows the supervised comparison. The physics residual remains near $0.93$ throughout, as expected of a privileged reference with access to the simulator's expected energy. Among deployable methods, the RF residual leads at low contamination, by a modest $0.02$--$0.03$ ROC-AUC, but its performance decays steeply as contamination grows, because the regression model increasingly absorbs injected inefficiency into its learned baseline. From $\rho = 20\%$ onward the proposed score overtakes the RF residual and retains the advantage for the remainder of the sweep, ending $0.05$ points ahead at $\rho = 99\%$ ($0.714$ versus $0.665$).
The proposed score also exceeds the LR and Huber residuals across most of the sweep.

Table~\ref{tab:ap} adds the PR-AUC view under class imbalance. Over the full sweep, RF attains a slightly higher mean PR-AUC ($0.741$ versus $0.713$), driven entirely by the low-contamination regime; beyond the crossover at $\rho = 25\%$ the proposed score achieves higher PR-AUC at every evaluated rate, leading in $8$ of the $14$ sampled scenarios. This trade-off matters operationally since the true prevalence of inefficiency is unknown before investigation, and training data for residual models are themselves contaminated to an unknown degree. A method that is competitive when inefficiency is rare and clearly stronger across the wide moderate-to-high contamination range provides the more robust prioritisation signal under this uncertainty.

\begin{table}[!ht]
\caption{PR-AUC Across the Contamination-Rate Sweep}
\label{tab:ap}
\centering
\footnotesize
\begin{tabular}{@{}lccc@{}}
\toprule
Contamination range & MDE (proposed) & RF residual & Stronger \\
\midrule
$1\%$--$25\%$   & 0.432 & \textbf{0.583} & RF \\
$26\%$--$99\%$  & \textbf{0.889} & 0.839 & MDE \\
Full sweep      & 0.713 & \textbf{0.741} & RF \\
\bottomrule
\end{tabular}
\end{table}

\subsection{Effect of the Embedding on Standard Detectors}

Table~\ref{tab:embedding} isolates the representational contribution at the reference contamination rate of $10\%$, with all methods scored on the full population of $5{,}000$ sites (values are therefore not directly comparable to the half-partition results in Fig.~\ref{fig:beta_sensitivity}). Moving from the original features to the learned embedding raises the ROC-AUC of Isolation Forest from $0.569$ to $0.906$, of the GMM from $0.577$ to $0.897$, and of the AE from $0.724$ to $0.889$: the energy-aware distortion objective reorganises the data into a space in which injected inefficiency is separable by generic detectors. The
proposed displacement score performs best overall (ROC-AUC $0.911$,
Precision@$10\% = 0.554$), indicating that the embedding is valuable in its own right and that the peer-relative scoring rule adds a further increment by aligning the score with the geometry the objective creates. LOF is the exception, performing better in the original space, a reminder that not every density-based detector benefits from a representation optimised for displacement.

\begin{table}[!ht]
\caption{Standard Detectors in the Original Feature Space Versus the Learned
MDE Embedding ($\rho = 10\%$)}
\label{tab:embedding}
\centering
\footnotesize
\setlength{\tabcolsep}{4pt}
\begin{tabular}{@{}lccc@{}}
\toprule
Method & ROC-AUC & PR-AUC & Prec.@10\% \\
\midrule
\textbf{MDE rel.\ displ.\ (proposed)} & \textbf{0.9105} & \textbf{0.5578} & \textbf{0.5540} \\
iForest (MDE embedding)   & 0.9055 & 0.5370 & 0.5200 \\
GMM (MDE embedding)       & 0.8972 & 0.4946 & 0.4900 \\
AE (MDE embedding)        & 0.8891 & 0.4878 & 0.4980 \\
LOF (features)            & 0.7830 & 0.3723 & 0.4060 \\
AE (features)             & 0.7242 & 0.3305 & 0.3300 \\
LOF (MDE embedding)       & 0.5781 & 0.1403 & 0.1760 \\
GMM (features)            & 0.5774 & 0.1632 & 0.1640 \\
iForest (features)        & 0.5685 & 0.1282 & 0.1620 \\
\bottomrule
\end{tabular}
\end{table}

\begin{figure}[ht]
\centering
\includegraphics[width=0.8\columnwidth]{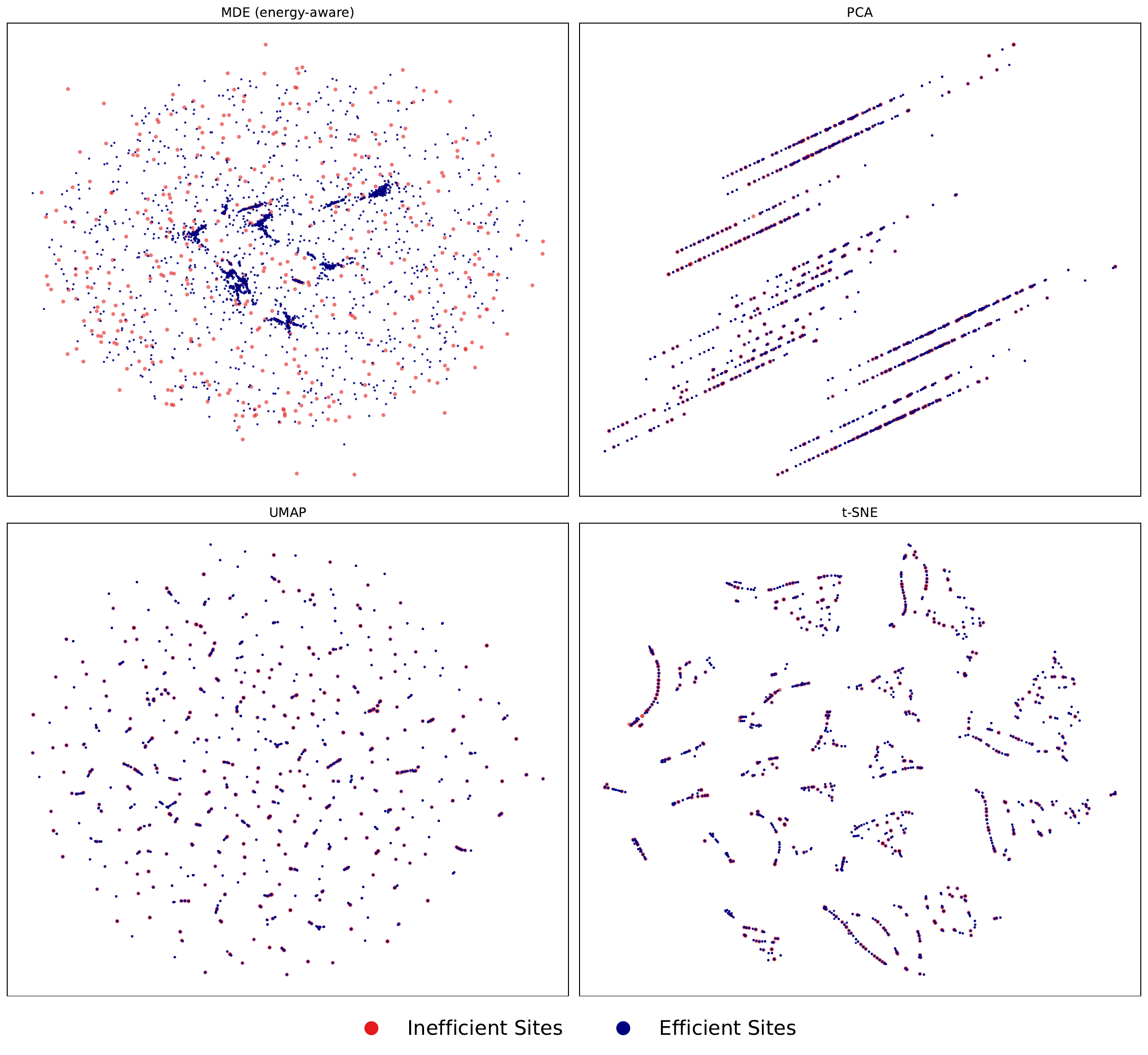}
\caption{Two-dimensional embeddings of the synthetic site population produced by the proposed energy-aware MDE objective, PCA, UMAP, and t-SNE. Red points denote injected inefficient sites; blue points denote efficient sites.}
\label{fig:embedding}
\end{figure}

Fig.~\ref{fig:embedding} complements the quantitative comparison with a
qualitative view against PCA, UMAP, and t-SNE. This comparison is intended solely for illustration: the MDE panel is generated from a dedicated two-dimensional embedding for visualisation, rather than the four-dimensional embedding used for anomaly scoring. The alternative embeddings organise the data by global variance or local density, leaving efficient and injected sites interspersed, whereas the energy-aware objective pushes injected sites toward peripheral regions, precisely the geometry the displacement score exploits.

\subsection{Teacher--Student Distillation}

Table~\ref{tab:distill} reports the distillation results at the reference setting; all test-set values are computed on the $1{,}500$-site held-out split, so they differ slightly from the full-population values in Table~\ref{tab:embedding}. On the held-out test set the teacher score achieves the highest ROC-AUC ($0.913$). The student models trained purely on its pseudo-labels retain most of this signal while improving inspection-oriented metrics: RF and XGBoost reach Precision@$10\% = 0.587$ (versus $0.567$ for the teacher), and XGBoost attains the best student PR-AUC ($0.642$). The geometric signal discovered by the embedding can therefore be transferred into simple, fast classifiers operating on raw site features which is attractive for deployment, where scoring new sites with a trained classifier is cheaper than
re-solving the embedding, and where feature-attribution tools can be applied to the student for interpretability.

\begin{table}[!t]
\caption{Teacher--Student Distillation Results ($\rho = 10\%$)}
\label{tab:distill}
\centering
\footnotesize
\setlength{\tabcolsep}{3pt}
\begin{tabular}{@{}llccc@{}}
\toprule
Method & Split & ROC-AUC & PR-AUC & P@10\% \\
\midrule
Teacher (MDE displacement) & Test & \textbf{0.9130} & 0.5878 & 0.5667 \\
\midrule
RF student        & Train & 0.9166 & 0.6640 & 0.5914 \\
                  & Test  & 0.8906 & 0.6088 & \textbf{0.5867} \\
XGBoost student   & Train & 0.9067 & 0.6829 & 0.6200 \\
                  & Test  & 0.8768 & \textbf{0.6415} & \textbf{0.5867} \\
Logistic student  & Train & 0.8503 & 0.5678 & 0.5143 \\
                  & Test  & 0.7973 & 0.5123 & 0.4600 \\
\bottomrule
\end{tabular}
\end{table}

\subsection{Discussion and Limitations}

Four findings summarise the study. First, peer-relative energy inconsistency is not discoverable by generic outlier detection in the raw feature space; it must be encoded into the representation, which the proposed push--pull objective does directly. Second, the framework's advantage over supervised residual modelling grows precisely where residual modelling is most fragile, when training data are more heavily contaminated by unlabelled inefficiency, which is the regime operators cannot rule out in practice. Third, the framework produces an interpretable geometric form of evidence: potential inefficiency is represented as the displacement of a site from its structural peers in the embedding, rather than as a residual from a fitted prediction model, and the two-dimensional comparison in Fig.~\ref{fig:embedding} shows that this geometry arises from the energy-aware objective itself rather than from generic dimensionality reduction. Fourth, the learned embedding has representational value beyond the proposed score, and its ranking can be distilled into lightweight classifiers, sketching a deployment pathway from unsupervised discovery to operational scoring.

The evaluation has several limitations. Ground truth is available only through controlled injection: although the injected inefficiencies are based on documented failure modes and calibrated to operational data, real inefficiencies may exhibit more complex temporal and environmental behaviour. The study is also cross-sectional, using one observation per site, and therefore does not capture seasonal variation, gradual equipment degradation, or maintenance events. Hyperparameters ($k_{\mathrm{graph}}$, $q$, and $\beta$) were selected once at the reference contamination rate (Section~\ref{sec:setup}) and held fixed across the contamination sweep, leaving adaptive hyperparameter selection as future work. In deployment, where injected labels are unavailable, the controlled-injection protocol of Section~\ref{sec:dataset} can itself serve as the tuning procedure: hyperparameters are selected against synthetic ground truth injected into the operator's own reference data before the framework scores the live network. Finally, sites with uncommon structural configurations may have only weakly comparable peers, making their peer-relative anomaly scores less reliable.

\section{Conclusion}
\label{sec:conclusion}

This paper developed and evaluated an unsupervised peer-relative framework for prioritising candidate energy-inefficient mobile network sites. Its central premise is that inefficiency cannot be judged from absolute consumption alone: sites differ substantially in structure, equipment, sharing status, vendor, and demand, so a site becomes a stronger candidate when its energy behaviour is inconsistent with structurally comparable peers. The proposed energy-aware MDE
formulation embeds this premise directly in a representation learning
objective, displacing energy-inconsistent sites from their neighbourhoods and scoring them by normalised displacement.

In a controlled evaluation environment built from operational data of
$5{,}372$ live sites, the proposed score outperformed standard unsupervised detectors at every contamination rate, remained competitive with supervised residual methods where they are strongest and clearly ahead where they are fragile, improved generic detectors when they operate in the learned embedding, and transferred its ranking to lightweight classifiers through pseudo-labels. The framework turns heterogeneous, unlabelled site data into an
actionable prioritisation signal for field investigations, supporting the industry's transition toward energy-efficient and environmentally responsible network operation.

Importantly, the framework has also demonstrated practical value during operational field investigations. Several high-ranking candidate sites identified by the framework were confirmed to contain previously undetected energy inefficiencies. At one Distributed Antenna System (DAS) site, an incorrect control configuration caused both air-conditioning units to operate continuously. Correcting the control logic to alternate between the two units reduced energy consumption by approximately 600~kWh per month. At a shopping centre site, both air-conditioning units and the free-cooling fan were operating simultaneously despite conditions where free cooling alone was sufficient. At another site adjacent to a grain silo, dust obstructing the ventilation inlets caused elevated internal temperatures. Cleaning the vents and increasing the allowable equipment temperature and air-conditioner setpoint restored efficient cooling. These examples illustrate that the proposed framework is capable of identifying operational inefficiencies that are not apparent from absolute energy consumption alone and can lead directly to measurable energy savings following field verification.

Future work will evaluate the framework on real operational deployments with partially validated labels, extend the controlled design with seasonal cooling, equipment degradation, and traffic--configuration interactions, incorporate temporal site trajectories, and develop the distillation pathway with confidence-weighted training and calibrated pseudo-label thresholds \cite{song2022noisy}, as well as hybrid residual--embedding formulations that combine predictive energy modelling with peer-relative geometric interpretation.

\PaperAppendix{Controlled Inefficiency Injection Specification}

Injected sites are selected by random permutation, independently of site configuration, and assigned one of four mechanisms in equal proportion. Let $e^{(0)}_i$ be the noisy baseline of \eqref{eq:noise} and $\sigma_i = \hat{e}_i\,\sigma_{\log,i}$ the site-level noise standard deviation on the energy (kWh) scale implied by \eqref{eq:noise}.

\emph{Type 1 (multiplicative overload):}
$e_i = e^{(0)}_i m_i$ with $m_i \sim \mathcal{U}(1.2, 1.8)$.

\emph{Type 2 (cooling overhead):} $e_i = e^{(0)}_i + u_i$ with
$u_i \sim \mathcal{U}(\ell_{m_i}, h_{m_i})$, where the bounds depend on the mast group of site $i$: tower $(200,400)$, disguised $(150,350)$, rooftop $(80,200)$, pole $(100,250)$, and other $(100,200)$\,kWh.

\emph{Type 3 (idle-RF load):} with $c_i$ cells, traffic $f_i$, and median traffic $\tilde{f}$, the idle factor is $\eta_i = \max(1 - f_i/\tilde{f},\, 0.1)$, the preliminary overhead $o^{\mathrm{RF}}_i = \max(c_i, 5)^2\, \eta_i a_i$ with $a_i \sim \mathcal{U}(0.5, 1.5)$, and
$e_i = e^{(0)}_i + \max\bigl(o^{\mathrm{RF}}_i,\, 2\sigma_i\bigr)$. The
quadratic cell term scales the overhead with radio configuration size, and the signal-to-noise floor $2\sigma_i$ is the only guard applied.

\emph{Type 4 (non-RAN parasitic load):} with $n_i$ non-RAN units as in \eqref{eq:baselinemodel},
$o^{\mathrm{NR}}_i = (n_i + 1)^2 v_i$ with $v_i \sim \mathcal{U}(20, 50)$, and $e_i = e^{(0)}_i + \max\bigl(o^{\mathrm{NR}}_i,\, 2\sigma_i\bigr)$.

Final simulated metered energy is rounded to two decimals.

\bibliographystyle{IEEEtran}
\bibliography{refs}

\end{document}